%% file: main.tex
\documentclass{article}

    \PassOptionsToPackage{numbers, compress, square,}{natbib}

\usepackage[preprint]{Styles/neurips_2024}

\usepackage[utf8]{inputenc} 
\usepackage[T1]{fontenc}    
\usepackage{hyperref}       
\usepackage{url}            
\usepackage{booktabs}       
\usepackage{amsfonts}       
\usepackage{nicefrac}       
\usepackage{microtype}      
\usepackage{xcolor}         

\usepackage{wrapfig}
\usepackage{adjustbox}
\usepackage{caption}
\usepackage{subcaption}
\usepackage{graphicx}
\usepackage{float}
\usepackage{tcolorbox}
\usepackage{array}
\newcolumntype{?}{!{\vrule width 2pt}}
\usepackage{wrapfig}

\newtcolorbox{mytextbox}[1][]{%
    arc=4pt, 
    colback=pink, 
    colframe=purple, 
    width=0.3\textwidth, 
    fontupper=\sffamily, 
    #1
}

\title{CapeX: Category-Agnostic Pose Estimation from Textual Point Explanation}

\author{%
  Matan Rusanovsky \hspace{5pt} Or Hirschorn \hspace{5pt} Shai Avidan\\
  \\
  Tel Aviv University\\
  \\
  \href{https://github.com/matanr/capex}{\color{blue}\fontfamily{cmss}\selectfont https://github.com/matanr/capex}
}

\begin{document}

\maketitle

\begin{abstract}
Conventional 2D pose estimation models are constrained by their design to specific object categories. This limits their applicability to predefined objects. To overcome these limitations, category-agnostic pose estimation (CAPE) emerged as a solution. CAPE aims to facilitate keypoint localization for diverse object categories using a unified model, which can generalize from minimal annotated support images.
Recent CAPE works have produced object poses based on arbitrary keypoint definitions annotated on a user-provided support image. Our work departs from conventional CAPE methods, which require a support image, by adopting a text-based approach instead of the support image. 
Specifically, we use a pose-graph, where nodes represent keypoints that are described with text. This representation takes advantage of the abstraction of text descriptions and the structure imposed by the graph.
Our approach effectively breaks symmetry, preserves structure, and improves occlusion handling.
We validate our novel approach using the MP-100 benchmark, a comprehensive dataset spanning over 100 categories and 18,000 images. Under a 1-shot setting, our solution achieves a notable performance boost of 1.07\%, establishing a new state-of-the-art for CAPE. Additionally, we enrich the dataset by providing text description annotations, further enhancing its utility for future research.
\end{abstract}

\begin{figure}
  \includegraphics[width=\textwidth]{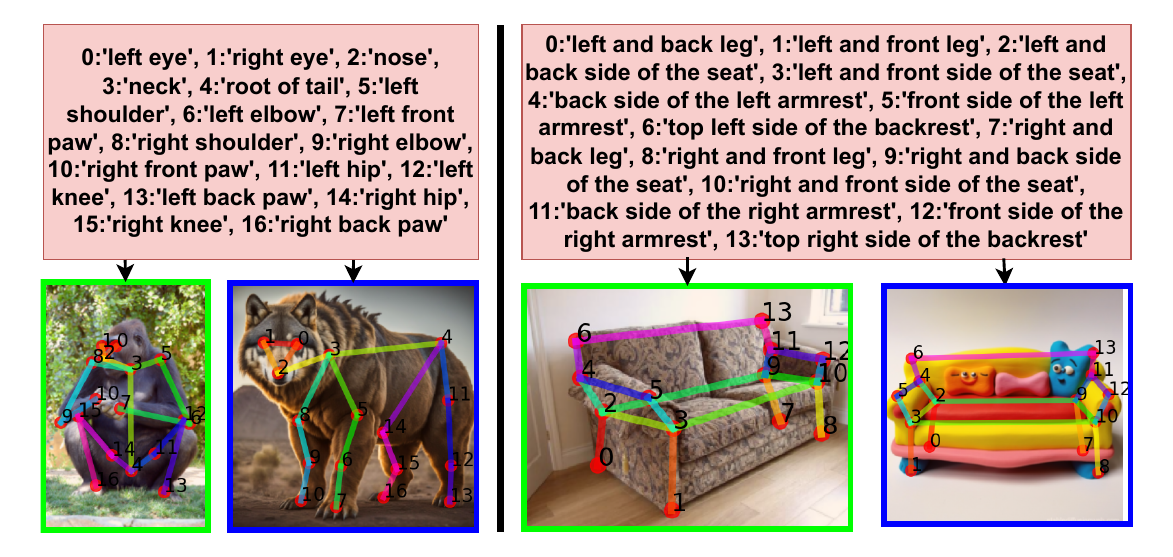}
  \caption{{\bf CapeX in action:} Given support keypoints text descriptions (in \textcolor{pink}{pink}) and a corresponding skeleton (not shown), our model localizes the skeleton on query images. In the first row, there are few input support text descriptions, and below each support input, there is a query image from the test set on the left (\textcolor{green}{green}), and an AI generated query image on the right (\textcolor{blue}{blue}). Our approach does not require a support image. Instead, it utilizes the abstraction power of text to improve keypoint localization.}
  \label{fig:teaser}
\end{figure}

\maketitle

\section{Introduction}
Pose estimation deals with the prediction of semantic parts' positions within objects depicted in images, a task crucial for applications like zoology, autonomous driving, virtual reality, and robotics~\cite{xu2022pose}. 
Previous pose estimation methods were typically constrained by their reliance on category-specific datasets for training. Consequently, when confronted with novel objects, these methods often exhibit limited efficacy due to their lack of adaptability. 

To address this challenge, recent research has introduced category-agnostic pose estimation (CAPE)~\cite{xu2022pose}, a paradigm capable of localizing semantic parts across diverse object categories, based on a single or few support examples. All previous CAPE works require a small set of support images annotated with the keypoints of interest. These support images are used in order to find the best spatial arrangement of the keypoints in the query image, based on latent visual correspondence to the annotated support keypoints.

This raises two challenges. First, the need to provide annotated support image(s) is cumbersome. Second, relying solely on visual correspondence between keypoints in different images, even from the same category, may lead to suboptimal results. This is because no two distinct images share parts with the exact same appearance. Still, both images should share parts with the same semantic meaning. For example, all cats have a head, legs, and a tail, but they never look the same. This idea is even more crucial when the objective is to estimate poses of objects in images from novel categories (i.e., dogs), as in CAPE.


We cope with these limitations by adopting a holistic approach to pose estimation, based on a support graph as input, with open-vocabulary textual descriptions on its nodes. No support images are needed. Instead of exclusively relying on visual support data, we leverage the abstraction power of textual data. This comprehensive view enables us to match the query keypoints' appearance to the textual description of the support keypoints, eliminating the need for support images altogether. 
Furthermore, following Pose Anything~\cite{hirschorn2023pose}, instead of treating the input keypoints as isolated entities, we treat them as structure-aware connected nodes of a graph.
By doing so, we effectively leverage the inherent relationships and dependencies between keypoints, enhancing the overall performance, breaking symmetry, preserving structure, and better handling occlusions. Figure \ref{fig:teaser} demonstrates our approach.

To evaluate the efficacy of our proposed method, we utilize the extended version~\cite{hirschorn2023pose} of the CAPE benchmark, MP-100~\cite{xu2022pose}. This dataset consists of more than 18,000 images spanning 100 categories, encompassing diverse subjects such as animals, vehicles, furniture, and clothes.
As some of the categories miss the keypoints' text descriptions, we collected and unified the text descriptions of the keypoints in all categories.
Our method is evaluated against previous CAPE methodologies. Notably, our approach surpasses the performance of existing methods, showcasing a new state-of-the-art performance under the 1-shot setting.

In summary, our contributions can be outlined as follows:
\begin{itemize}
    \item We propose modeling the support keypoints using connected graph nodes coupled with text descriptions as opposed to previous methods that rely on visual signals. This methodology matches the support to the query keypoints, thanks to the abstraction power of text and graphs. Furthermore, this approach does not require support images for either training or inference.
    \item We provide an enhanced version of the MP-100 dataset with textual annotations for the keypoints in all categories, enriching the benchmarking capabilities for category-agnostic pose estimation.
    \item We establish new benchmarks in category-agnostic pose estimation, showcasing state-of-the-art performance on the MP-100 dataset, without finetuning the support feature extraction.
\end{itemize}

\section{Related Work}
\subsection{Category-Agnostic Pose Estimation}
The primary aim of pose estimation is to localize the semantic keypoints of objects or instances precisely. Traditionally, pose estimation methods have been largely tailored to specific categories, such as humans~\cite{alphapose, openpose, yang2021transpose}, animals~\cite{yu2021ap, yang2022apt}, or vehicles~\cite{song2019apollocar3d, reddy2018carfusion}. However, these prior works are constrained to object categories encountered during training.

An emerging aspect in this field is category-independent pose estimation, as introduced by POMNet~\cite{xu2022pose}. This few-shot approach predicts keypoints by comparing support keypoints with query images in the embedding space, addressing the challenge of object categories not seen during training.
POMNet employs a transformer to encode the support keypoints and query images. It uses a regression head to predict similarity from the extracted features. 
CapeFormer~\cite{shi2023matching} extends this matching paradigm to a two-stage framework, refining unreliable matching outcomes to improve prediction precision.
Pose Anything~\cite{hirschorn2023pose} presented a significant departure from previous CAPE methods, which refer to keypoints as isolated entities, by treating the input pose data as a graph. 
It utilizes Graph Convolutional Networks (GCNs) to leverage the inherent object's structure to break symmetry, preserve the structure, and better handle occlusions. However, similar to previous CAPE models, it relies solely on visual features.
Our work builds upon Pose Anything, utilizing its structure-aware architecture, while introducing the abstraction power of text.

\subsection{Open-Vocabulary Models}
A growing area in computer vision called Open-Vocabulary learning is being explored in various vision tasks. These new methods aim to localize and recognize categories beyond the labeled space. The open-vocabulary approach is broader, more practical, and more efficient compared to weakly supervised setups~\cite{wu2024towards}.
Large-scale vision-language models (VLMs) like CLIP~\cite{radford2021learning} and ALIGN~\cite{jia2021scaling} have shown promise in handling both visual and text data, and proved useful for open-vocabulary tasks.

Open-vocabulary object detection (OVOD) using VLMs was utilized by performing object-centric alignment of language embeddings from the CLIP model~\cite{bangalath2022bridging}. Zang et al.~\cite{zang2022open} suggested a DETR (common transformer-based architecture) based detector, able to detect any object given its class name. In addition, LLMs were also used to generate informative language descriptions for object classes and construct powerful text-based annotations~\cite{kaul2023multi}.
Another task that recently achieved significant progress is open-vocabulary semantic segmentation (OVSS), which aims to segment objects with arbitrary text. One line of research~\cite{ding2022decoupling,xu2022simple, xu2023open} combines powerful segmentation models like MaskFormer~\cite{cheng2021per} with CLIP~\cite{Xu_2023_CVPR} while others~\cite{zou2024segment} utilize foundation segmentation models like SAM~\cite{kirillov2023segment}.
Recently, Wei et al.~\cite{wei2024ov} suggested a new benchmark for Open-Vocabulary Part Segmentation, to further enhance open-vocabulary capabilities.

Yet, there's still limited exploration into open-vocabulary keypoint detection.
Recently, CLAMP~\cite{zhang2023clamp} leveraged CLIP to prompt animal keypoints. They found that establishing effective connections between pre-trained language models and visual animal keypoints is challenging due to the substantial disparity between text-based descriptions and keypoint visual features. CLAMP attempts to narrow this gap by using contrastive learning to align the text prompts with the animal keypoints during training.
Our approach aims for general keypoint estimation of any category while taking advantage of structure as a prior for localization by treating the input prompts as a graph.


\section{Method}\label{sec:method}

\subsection{Open-Vocabulary Keypoint Detection}

\begin{figure}
\centering
     \begin{subfigure}[b]{0.3\textwidth}
         \centering
         \includegraphics[width=\textwidth]{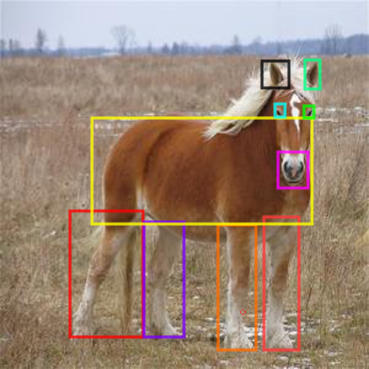}
         \caption{}
         \label{fig:horse_detection}
     \end{subfigure}
          \begin{subfigure}[b]{0.3\textwidth}
         \centering
         \includegraphics[width=\textwidth]{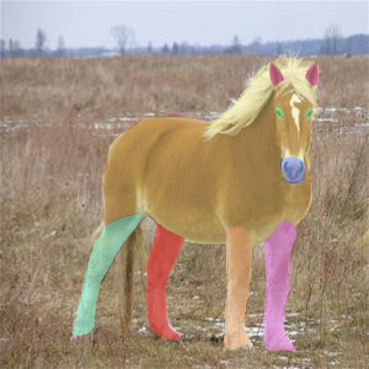}
         \caption{}
         \label{fig:horse_seg}
     \end{subfigure}
          \begin{subfigure}[b]{0.3\textwidth}
         \centering
         \includegraphics[width=\textwidth]{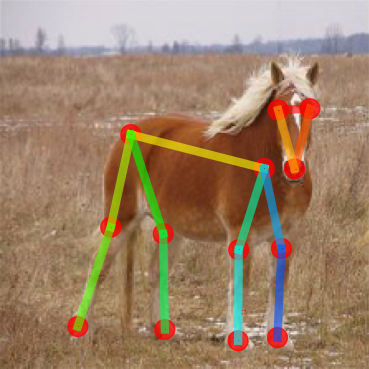}
         \caption{}
         \label{fig:horse_pose}
     \end{subfigure}
     \caption{\textbf{Different Open-Vocabulary Tasks:} We show three different open-vocabulary tasks: (a) object detection, (b) part segmentation, and (c) keypoint detection. Object detection identifies objects and locations, segmentation provides pixel-level semantic details, and keypoint detection offers finer localization than object detection while being more practical for localization than segmentation.
     }
  \label{fig:open_vocabulary}
\end{figure}

Open-vocabulary learning seeks to localize and recognize categories beyond those included in annotated labels. While open-vocabulary object detection and segmentation have gained attraction, keypoint detection has largely been overlooked. 
Segmentation offers pixel-level details about semantic regions, whereas object detection identifies specific objects and their locations. Keypoint detection lies between these two, offering finer semantic localization than object detection, yet being more lightweight and practical for parts localization compared to segmentation. Figure ~\ref{fig:open_vocabulary} demonstrates the differences between the three tasks.
\begin{figure}
  \includegraphics[width=\textwidth]{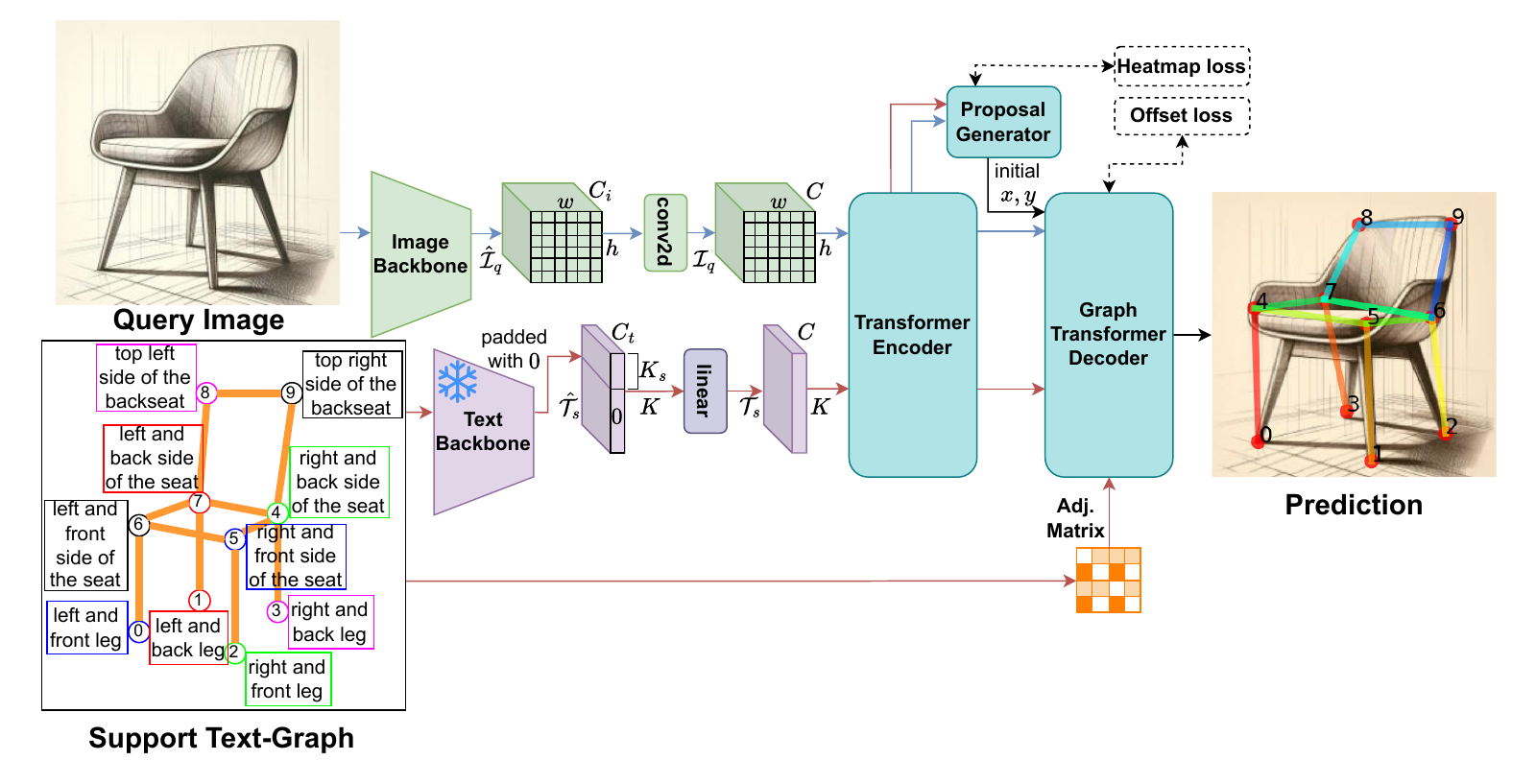}
  \caption{{\bf Architecture overview:} Our framework uses image and text backbones benefiting from both accurate and abstract descriptions respectively. The extracted feature descriptors are forwarded into the transformer encoder that refines them. The refined features are passed into the proposal generator alongside the graph transformer decoder, utilizing the graph structure within the data.
  } 
  \label{fig:overview}
\end{figure}

Open-vocabulary keypoint detection aims to use natural language to identify any keypoints in images, even if those key categories were not part of the training data. Advances in vision-language models such as CLIP allow keypoint detectors to harness powerful language models to perform language-driven tasks.
We introduce a new open-vocabulary keypoint detector inspired by CAPE, a few-shot task of localizing keypoints in unseen categories using a few annotated images.
The core idea of our work is that for the task of CAPE, it is more beneficial to describe the searched points in the query image using text description instead of relying only on the visual features of the support images. This is because text allows a higher level of abstraction and offers a looser restriction to the support request. This is true even when the support and query images are from the same category - for example, no two cats share visually the exact same \textit{front left leg}, but both cats have a part within them that follows the same text description: \textit{front left leg}. This distinction is even more significant when dealing with images from different categories as in CAPE. We present in the supplemental Figure~\ref{fig:same_query} how a support image-based CAPE solution might suffer from incorrect pose estimation due to visually inconsistent support images. 
Bottom left in Figure~\ref{fig:overview} is an example of a support text-graph that our system utilizes.

\subsection{Text Prompts as Visual Queues}\label{sec:architecture}

Our framework extracts visual features from the query image and matches them to the textual features that are extracted from the support text-graph. We incorporated this notion by introducing text comprehension into Pose Anything's framework \cite{hirschorn2023pose}.

A pre-trained and fine-tuned SwinV2-S~\cite{liu2022swin} is utilized for extracting image features from the input query image producing the feature map $\hat{\mathcal{I}}_q \in \mathbb{R}^{hw \times C_i}$, where $hw$ is the total number of patches and $C_i$ is the image embedding dimension. Then $\hat{\mathcal{I}}_q$ is passed through a 1x1 convolutional layer, resulting in $\mathcal{I}_q \in \mathbb{R}^{hw \times C}$.

The support keypoint text descriptions are embedded in our model using a pre-trained gte-base-v1.5~\cite{li2023towards}. The text embeddings of all $K_s$ keypoints of the provided support sample are then normalized. The normalized keypoints are padded with zeros, effectively resulting in $K$ keypoints, where $K$ is defined to be the maximum amount of possible keypoints in the dataset. The final text feature map is of the form $\hat{\mathcal{T}}_s \in \mathbb{R}^{K \times C_t}$ where $C_t$ is the text embedding dimension. Then $\hat{\mathcal{T}}_s$ is passed through a linear layer resulting in $\mathcal{T}_s \in \mathbb{R}^{K \times C}$. 
During training, the text backbone is frozen.
This approach also offers a lighter optimization procedure, as the gradients of the text features are ignored. An architecture overview is presented in Figure \ref{fig:overview}.

The extracted query image features and the support descriptions features are then refined using the transformer encoder. This encoder comprises three transformer blocks. Since the embedding spaces of the support text and query image differs, the support and query features are first fused together and then separated again. This practice aids in closing the gap between their representations \cite{shi2023matching} using self-attention layers. 
Then, similarity heatmaps between the query and support features are formed, using the proposal generator. The proposal generator utilizes a trainable inner-product mechanism~\cite{shi2022represent} to explicitly represent similarity. Peaks are then chosen from these maps to act as the basis for similarity-aware proposals. 
A graph transformer decoder network receives these initial proposals, processes them using a combination of attention and Graph Convolutional Network (GCN) layers, and predicts the final estimated keypoints locations. 
Utilizing GCN layers allows for the explicit consideration of semantic connections between keypoints, thereby benefiting CAPE tasks. 
We visualize cross-attention maps from the decoder trained with text prompts compared to visual prompts in the supplemental (Figure \ref{fig:attentions}).

To train our end-to-end method we use two loss terms: \(\mathcal{L}_{heatmap}\) and \(\mathcal{L}_{offset}\). The former penalizes the similarity metric while the latter penalizes the localization output:
    \begin{equation}
        \mathcal{L}_{heatmap} = \frac{1}{(K \cdot H \cdot W)}\sum_{i=1}^{K}{||\sigma(M_i)-H_i||}
    \end{equation}
    \begin{equation}
        \mathcal{L}_{offset} = \frac{1}{L}\sum_{i=1}^{L}{\sum_{i=1}^{K}{|P_i^l-\hat{P_i}|}}
    \end{equation}
    where \(\sigma\) is the sigmoid function, and for each point \(i\),  \(M_i\) is the output similarity heatmap of the proposal generator, \(H_i\) is the ground truth heatmap, \(P_i^l\) is the output location from layer \(l\) and \(\hat{P_i}\) is the ground truth location. The overall loss is:
    \begin{equation}
            \mathcal{L} = \lambda_{heatmap} \cdot \mathcal{L}_{heatmap} + \mathcal{L}_{offset}
        \end{equation}

\section{Experiments}\label{sec:experiments}
\input{Tables/benchmark}

In line with prior CAPE studies, we utilize the MP-100 dataset~\cite{xu2022pose} as both our training and evaluation dataset, which comprises samples sourced from existing category-specific pose estimation datasets~\cite{lin2014microsoft, sagonas2016300, koestinger2011annotated, wang2018mask, ge2019deepfashion2, yu2021ap, labuguen2021macaquepose, pereira2019fast, graving2019deepposekit, welinder2010caltech, reddy2018carfusion, khan2020animalweb, wu2016single}. This dataset consists of over 18K images spread across 100 distinct sub-categories and 8 super-categories (human hand \& face \& body, animal face \& body, clothes, furniture and vehicle), featuring varying numbers of keypoints, ranging from 8 to 68 keypoints.

The dataset is divided into five separate splits for training and evaluation. Importantly, each split ensures that the categories used for training, validation, and testing are mutually exclusive, ensuring that the categories used for evaluation are unseen during the training phase.

The original dataset comes with partial skeleton annotations in different formats, including variations in the keypoint indexing. We use the updated version of Pose Anything~\cite{hirschorn2023pose} that includes unified skeleton definitions for all categories. The updated version predominantly featured brief text sentences describing each point within most categories. However, certain categories exhibited text descriptions with distinct characteristics, such as the use of underscores between words instead of spaces, while others lacked any text descriptions altogether.
We annotated and standardized the text descriptions of all points in all categories, offering a new supervision capability to the updated version of~\cite{hirschorn2023pose} of the original MP-100. 



To assess our model's performance, we employ the Probability of Correct Keypoint (PCK) metric~\cite{yang2012articulated}, setting a PCK threshold of 0.2, following the conventions established by Pose Anything~\cite{hirschorn2023pose}, POMNet~\cite{xu2022pose} and CapeFormer~\cite{shi2023matching}.
More design choices and evaluations are in the supplementary.

\paragraph{Implementation Details}\label{sec:implementation}
To ensure a fair comparison, except for the text backbone, the configuration settings remain consistent with Pose Anything~\cite{hirschorn2023pose} and CapeFormer~\cite{shi2023matching}. The trainable features of the framework remain exactly the same as in Pose Anything (except for the new linear layer) since the text backbone is frozen during training in our framework. However, we also evaluate and present the performance of the framework with an unfrozen text backbone in the supplemental Table \ref{tab:ablation_finetune}. $C_i$ is 768 in SwinV2-S, $C_t$ is 768 in gte-base-v1.5. $C$ and $K$ are set to $256$ and $100$, respectively.
The architecture is implemented within the MMPose framework~\cite{mmpose2020}, trained using the Adam optimizer for 200 epochs with a batch size of 16. The initial learning rate is $10^{-5}$, reducing by a factor of 10 at the 160th and 180th epochs. All experiments in our work were carried out using a machine equipped with an NVIDIA RTX A5000 GPU. Our model required $10$ GB of GPU memory and took roughly $20$ hours to train for each split.
\subsection{Benchmark Results}
\begin{figure}
  \includegraphics[width=\textwidth]{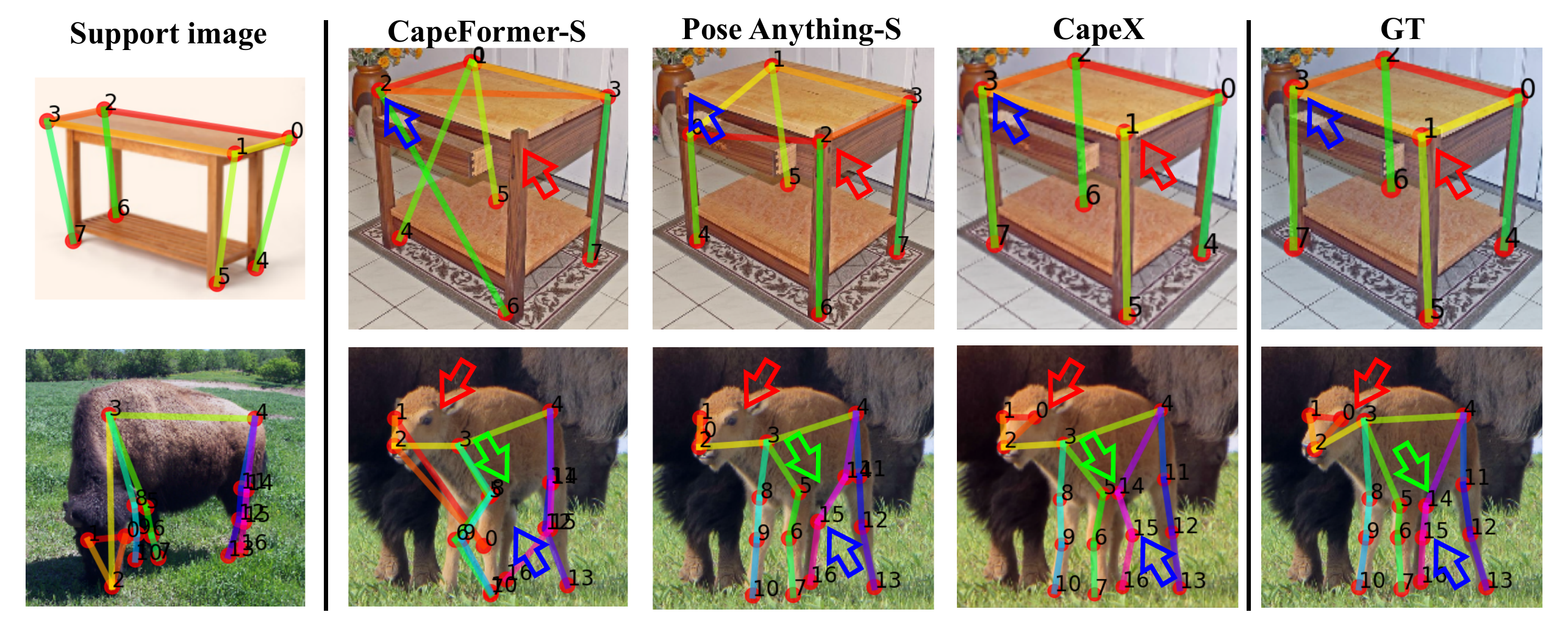}
  \caption{{\bf Qualitative results:} From left to right: support images that are used by the competitors, CapeFormer-S, Pose Anything-S, our model, and the GT. Support text descriptions used by our model are not shown. Main differences are pointed out using arrows.
  } 
  \label{fig:qualitative}
\end{figure}
We conduct a comparative analysis of our approach with gte-base-v1.5~\cite{li2023towards} as the freezed text backbone, against Pose Anything~\cite{hirschorn2023pose}, as well as prior CAPE methodologies such as CapeFormer~\cite{shi2023matching} and its enhanced version CapeFormer-T from~\cite{hirschorn2023pose}, POMNet~\cite{xu2022pose}, ProtoNet~\cite{snell2017prototypical}, MAML~\cite{finn2017model}, and Fine-tuned~\cite{nakamura2019revisiting}. For a comprehensive understanding of these models' performance, additional details can be found in~\cite{xu2022pose}.

Our evaluation is based on the MP-100 dataset, considering the 1-shot scenario. While traditionally 1-shot refers to a single required support image, our framework uses a single text-graph instead. We do not report the 5-shot results, because we do not use 5 different support images. The results are presented in Table \ref{tab:mp100}. Notably, our text-based approach outperforms Pose Anything on most splits, with an average improvement of 1.07\% under the 1-shot setting. These results establish a new state-of-the-art result, showcasing the efficacy of utilizing text-graphs for CAPE.


A qualitative comparison of our model against CapeFormer-S and Pose Anything-S in presented in Figure \ref{fig:qualitative}. Our model performs well given the support text-graph input (not shown), while the support image-based techniques are sensitive to the inconsistencies between the
support and query images. 

\subsection{Ablation Study}\label{sec:ablation}
\paragraph{Text Modifications}
\begin{figure}
    \centering
    \includegraphics[width=\linewidth]{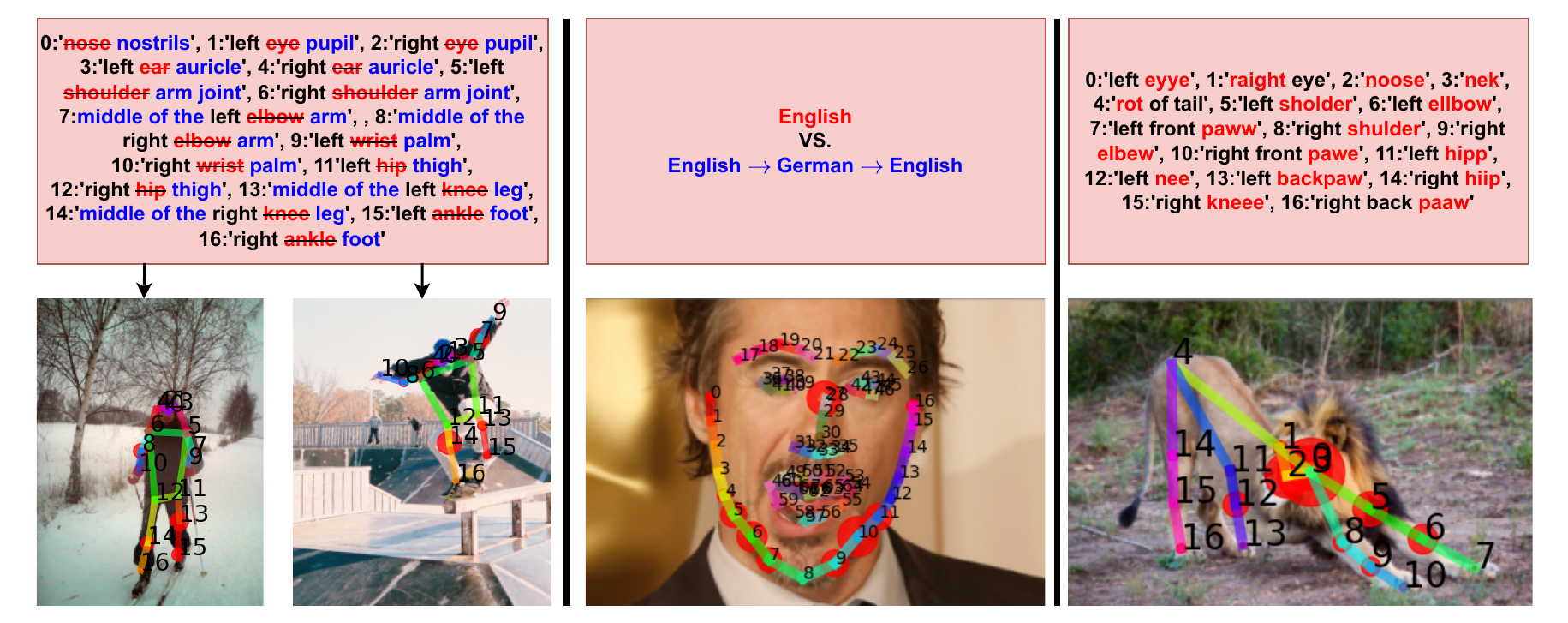}
    \caption{{\bf Modified text descriptions:} Top is the support keypoints text descriptions.
    Left is a {\bf synonym} words test, middle is a {\bf translation} test and right is {\bf typo} test. Below each description, query output(s) are presented. Each node in the presented graph is the average positions between the original and modified text descriptions. The diameter represents the distance between the positions.}
    \label{fig:synonyms}
\end{figure}
The fact that the text backbone was not fine-tuned during the training of our model, keeps it from overfitting to text descriptions from the training set. On the contrary, the model demonstrates its effectiveness across modified text inputs, while preserving similar estimated poses overall. We test the robustness of our model on different types of modifications for the keypoint descriptions in Figure \ref{fig:synonyms}. Specifically, we test the adaptability of the model to synonym descriptions (left), to translation to another language and back to English (middle), and to typos (right). 
Notably, all average keypoints are placed in acceptable positions.
The main differences in the two synonym test examples are in keypoints 7, 8 ('elbow' $\rightarrow$ 'middle of the arm') and 13, 14 ('knee' $\rightarrow$ 'middle of the leg'). In the translation test, the main differences are in keypoints 5-7 and 9-11 ('top/bottom side of the right/left jaw/cheek' $\rightarrow$ 'upper/lower side of the right/left jaw/cheek') and 27 ('top side of the nose' $\rightarrow$ 'upper side of the nose'). In the typos test, the significant inconsistencies are in the head (keypoints 0,2 and 3), while minor differences are spotted also in the leg joints (keypoints 5,6 and 12).
All these differences are compatible with the discrepancy imposed by the different descriptions.
\begin{figure}[h!]
    \centering
    \includegraphics[width=\linewidth]{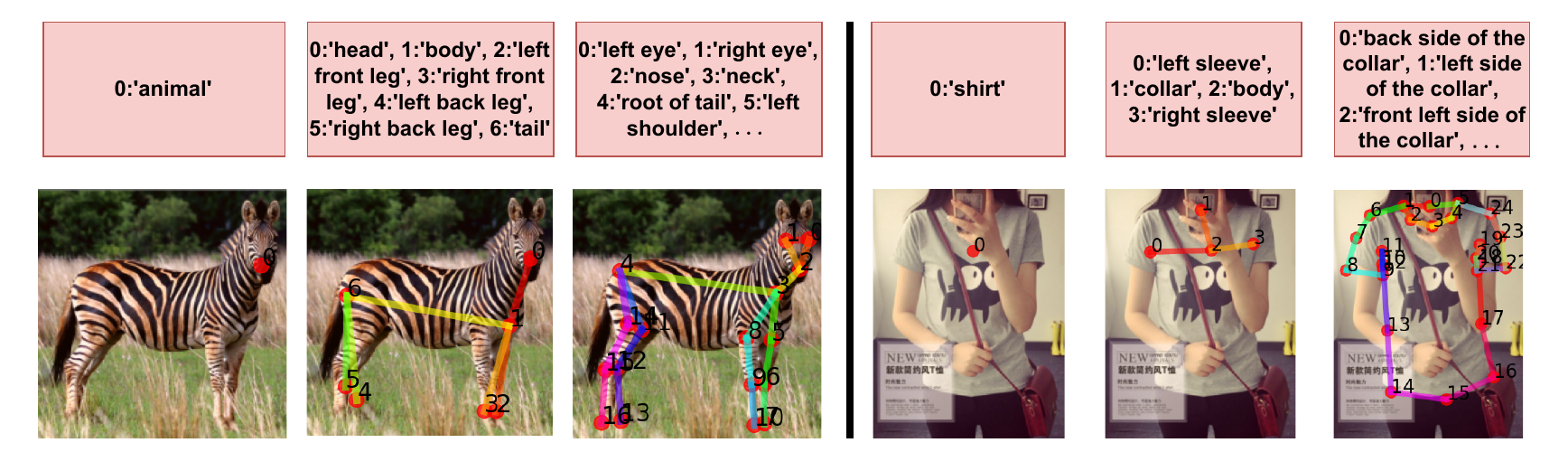}
    \caption{{\bf Text Abstractions:} Model performance over different levels of text-pose abstractions.}
    \label{fig:abstractions}
\end{figure}
\paragraph{Occlusions and Levels of Abstraction}
We test the robustness of our model on keypoints that are described using different levels of text and pose-graphs abstractions. Results are in Figure \ref{fig:abstractions}. Although not trained with the prompted text descriptions and pose-graphs, the model presents satisfactory results in both examples.

\begin{wrapfigure}{R}{0.6\textwidth}
    \centering
    \includegraphics[width=0.6\textwidth]{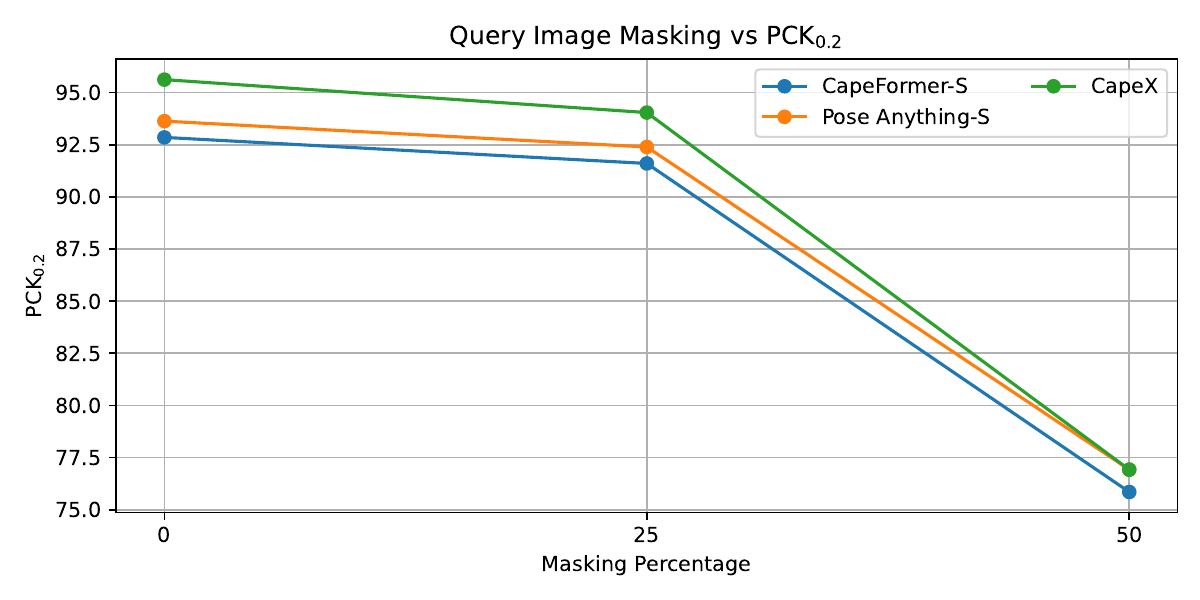}
    \caption{{\bf Masking the query image:} PCK$_{0.2}$ performance as a function of the masking percentage.}
    \label{fig:mask}
\end{wrapfigure}
Furthermore, we assess the effectiveness of text-graphs in handling occlusions within query images by applying random masks to them before estimating the support keypoints. Quantitative results are in Figure \ref{fig:mask}, and a qualitative comparison is presented in the supplemental Figure \ref{fig:mask_example}. Our method demonstrates superior performance over Pose Anything-S and CapeFormer-S in the entire presented occlusion range while maintaining similar degradation levels between 0\% to 25\%. The improved performance at lower masking percentages can be attributed to the text-graphs' abstraction capability and their ability to estimate missing keypoints relative to the visible ones. However, as our approach does not utilize a support image as input, performance significantly drops and matches the competitors, when a substantial portion of the image is occluded (50\%). This is because the absence of the query image leaves the model with insufficient information to operate effectively. This stands in contrast to traditional CAPE methodologies that incorporate a support image, which provides crucial structural cues. In such frameworks, the support image aids the model in hallucinating and extrapolating matching keypoints within the query, particularly when considering graph structures as in Pose Anything.

\paragraph{Out of Distribution Query Images}
\begin{figure}
    \centering
    \includegraphics[width=\textwidth]{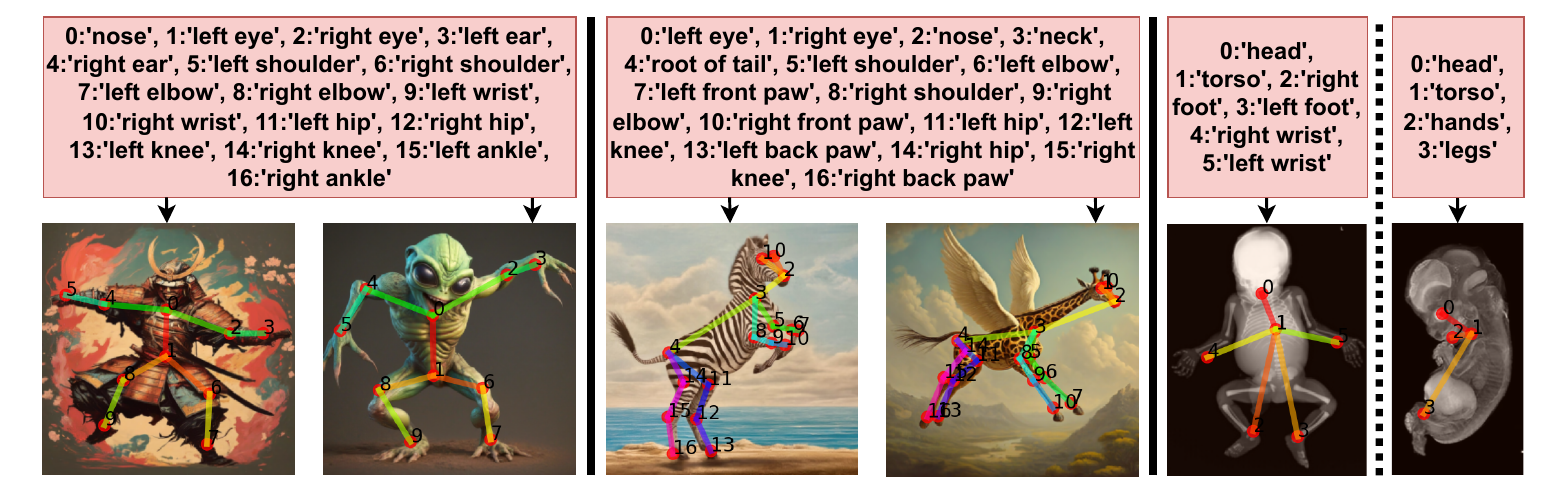}
    \caption{{\bf Out of distribution performance:} Top is the support keypoints text descriptions. Below each description, we present the query output.
    }
    \label{fig:OOD}
\end{figure}
We evaluate the resilience of our model to out-of-distribution query images generated via diffusion models. In Figure \ref{fig:OOD}, we examine novel styles, categories, poses, and even imaging methods. While the estimated poses generally align coherently with both the query and the support text-graph, there are notable inconsistencies. For example, the model appears to inaccurately localize 'knees' in both two left query outputs and fails to localize the 'front paws' in the zebra query output.
\section{Limitations}\label{sec:limitations}
\begin{figure}
    \centering
    \includegraphics[width=\textwidth]{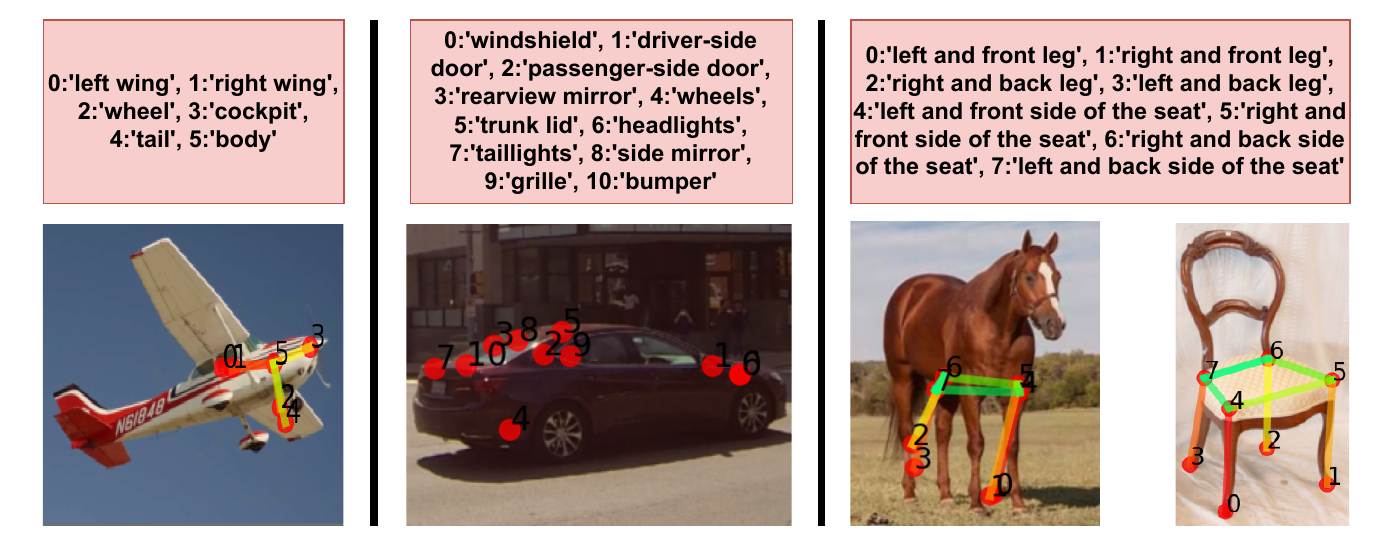}
    \caption{{\bf Failure cases:} From left to right: a category outside of the dataset, introducing vastly new keypoint descriptions, and cross-category descriptions. 
    }
    \label{fig:limitations}
\end{figure}
We stress that although our model strives for full open-vocabulary performance, it is still trained on a relatively small training set over arguably a short time period, compared to the state-of-the-art large vision-language foundation models. We present in Figure \ref{fig:limitations} a few failure cases that may be addressed in future research. Our model does not handle new categories with novel text-graphs well, as can be seen in the plane example on the left. In addition, prompting with vastly new parts may lead to incorrect localizations as can be seen in the car example on the middle (for example, driver/passenger-side door). Lastly, the model incorrectly executes semantically challenging descriptions. For example, the model can not localize a 'seat' in a horse, even though riders may seat on it. Instead, it hallucinates a pose of a chair that it has seen in the training set.

\section{Broader impact}\label{sec:broader}
Advancements in pose estimation technology can revolutionize fields such as autonomous driving, smart cities, sports, etc., by enabling precise movement analysis. However, when a pose estimation tool is used, specifically in fields such as surveillance, it is crucial to address privacy concerns and establish ethical guidelines to protect sensitive personal data and ensure responsible use.

\section{Conclusions}

CapeX is a Category Agnostic Pose Estimation (CAPE) approach that is based on text input. In particular, CapeX takes a pose-graph, where text descriptions are attached to its nodes, and finds these keypoints in a query image. This stands in contrast to previous CAPE approaches that require support image with annotated pose-graph as part of the input. CapeX can be viewed as an Open Vocabulary Kepoint Detection algorithm, closing the gap between Open Vocabulary Object Detection and Open Vocabulary Segmentation.

CapeX was tested on the standard MP-100 dataset and achieves a new state of the art result, surpassing previous CAPE methods that rely on support image as input instead of text.

\begin{ack}
This work was partly funded by the Weinstein Institute.
\end{ack}






\bibliographystyle{Styles/Reference-Format}
\bibliography{bibliography}


\newpage
\appendix

\section{Appendix / Supplemental Material}
\subsection{Disadvantages of Visual Prompts in CAPE}
\input{Figures/same_query/same_query}

\begin{figure}[h!]
\centering
     \begin{subfigure}[b]{0.3\textwidth}
         \centering
         \includegraphics[width=\textwidth, height=\textwidth]{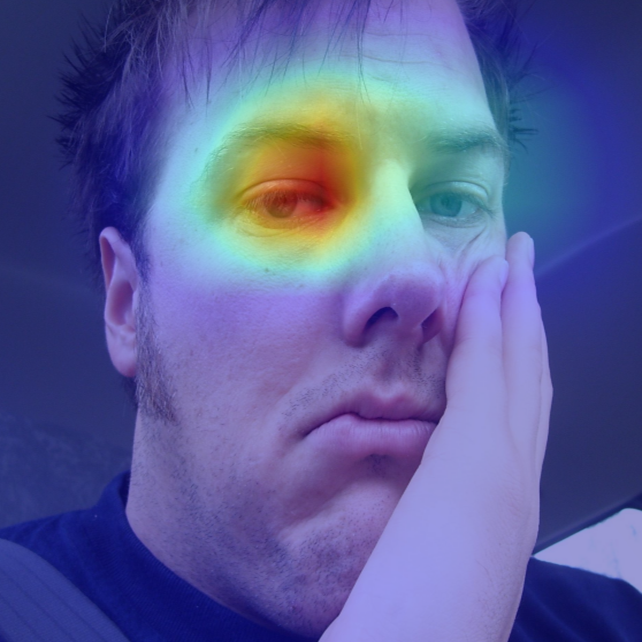}
         \caption{}
         \label{fig:y equals x}
     \end{subfigure}
     \hspace{0.5cm}
     \begin{subfigure}[b]{0.3\textwidth}
         \centering
         \includegraphics[width=\textwidth, height=\textwidth]{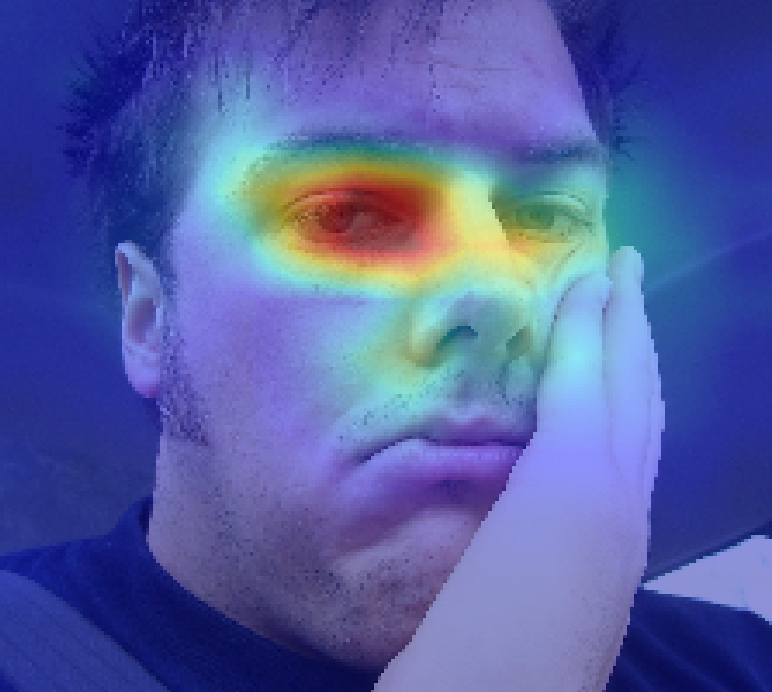}
         \caption{}
         \label{fig:y equals x}
     \end{subfigure}
      
  \caption{\textbf{Cross-attention maps:} comparison between the query image and the 'top right side of the left eye' keypoint. (a) is our model
  , and (b) is PoseAnything-S. Our model demonstrates in (a) better performance at breaking symmetry and distinguishing between left and right, compared to Pose Anything-S in (b), that attends more to the right eye and the nose. 
  }
  \label{fig:attentions}
\end{figure}
We exemplify the key disadvantage of support image-based CAPE approaches in Figure~\ref{fig:same_query}. Specifically, Pose Anything-S, suffers from incorrect pose estimations when prompted with visually inconsistent support images.

We also compare our models with regards to localization and symmetry breaking. We visualize cross-attention maps from the decoder trained with text prompts compared to visual prompts in Figure \ref{fig:attentions}. Our model breaks symmetry and distinguishes better between left and right, compared to Pose Anything. This can be seen by the lower attention to the right eye, when prompted with the keypoint 'top right side of the left eye'.
\subsection{Additional Experiments}
\subsubsection{Different Architectures}\label{sec:different}
\input{Tables/ablation_finetune}
We assess the framework's performance with or without fine-tuning applied to the text backbone. We explore two potential text backbones: gte-base-v1.5~\cite{li2023towards} and CLIP ViT-B/32~\cite{radford2021learning}. 
The optimal configuration appears to be the frozen gte-base-v1.5 as the text backbone, yielding superior performance. Interestingly, although gte-base-v1.5 boasts approximately 139 million trainable parameters compared to the 63 million parameters in the text module of CLIP ViT-B/32 (totaling 150 million parameters), training with either as a frozen text backbone consumed similar execution times, lasting roughly $20$ hours. Memory usage for loading both models required a similar volume of $10$ GB. 
However, fine-tuning both models incurred substantial costs in terms of memory: $15$ in gte and $30$ GB in CLIP, as well as in execution time: $35$ hours for both architectures, without yielding any performance improvements in both text backbones. 
The drop in performance can possibly be attributed to the fact that the text backbones suffer from overfitting during their tuning. This is somewhat expected as language models usually train on larger datasets over longer training sessions. 

We also tested the original MLP transformer decoder architecture as in~\cite{shi2023matching} with the best performing setting. Memory consumption and execution time using this transformer decoder were comparable to the graph transformer decoder. We find that utilizing the graph structure via the graph transformer decoder as in~\cite{hirschorn2023pose} slightly boosts the performance. Full results are presented in Table \ref{tab:ablation_finetune}.

\begin{figure}[h!]
    \centering
    \includegraphics[width=\linewidth]{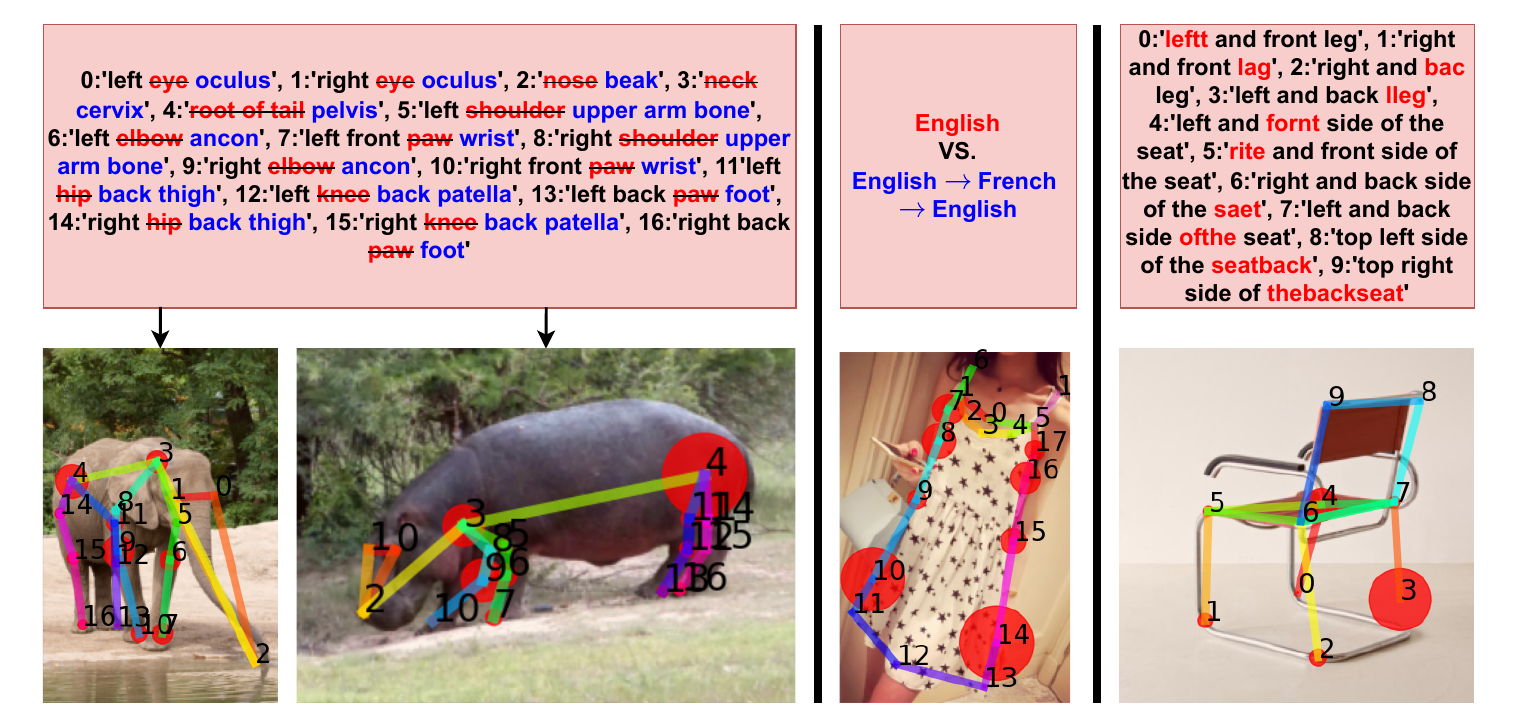}
    \caption{\textbf{Modified text descriptions:} Top is the support keypoints text descriptions.
    Left is a synonym words test, middle is a translation test and right is typo test. Below each description, query output(s) are presented. Each node in the presented graph is the average positions of the original and modified text descriptions. The diameter represents the distance between the positions.}
    \label{fig:synonyms2}
\end{figure}

\subsubsection{Adaptability to Support Text Modifications}

We provide additional examples of the ability of our model to adapt to different types of text modifications in Figure \ref{fig:synonyms2}.

\subsubsection{Occlusions and levels of abstraction}
\begin{figure}[h!]
    \centering
    \includegraphics[width=0.8\linewidth]{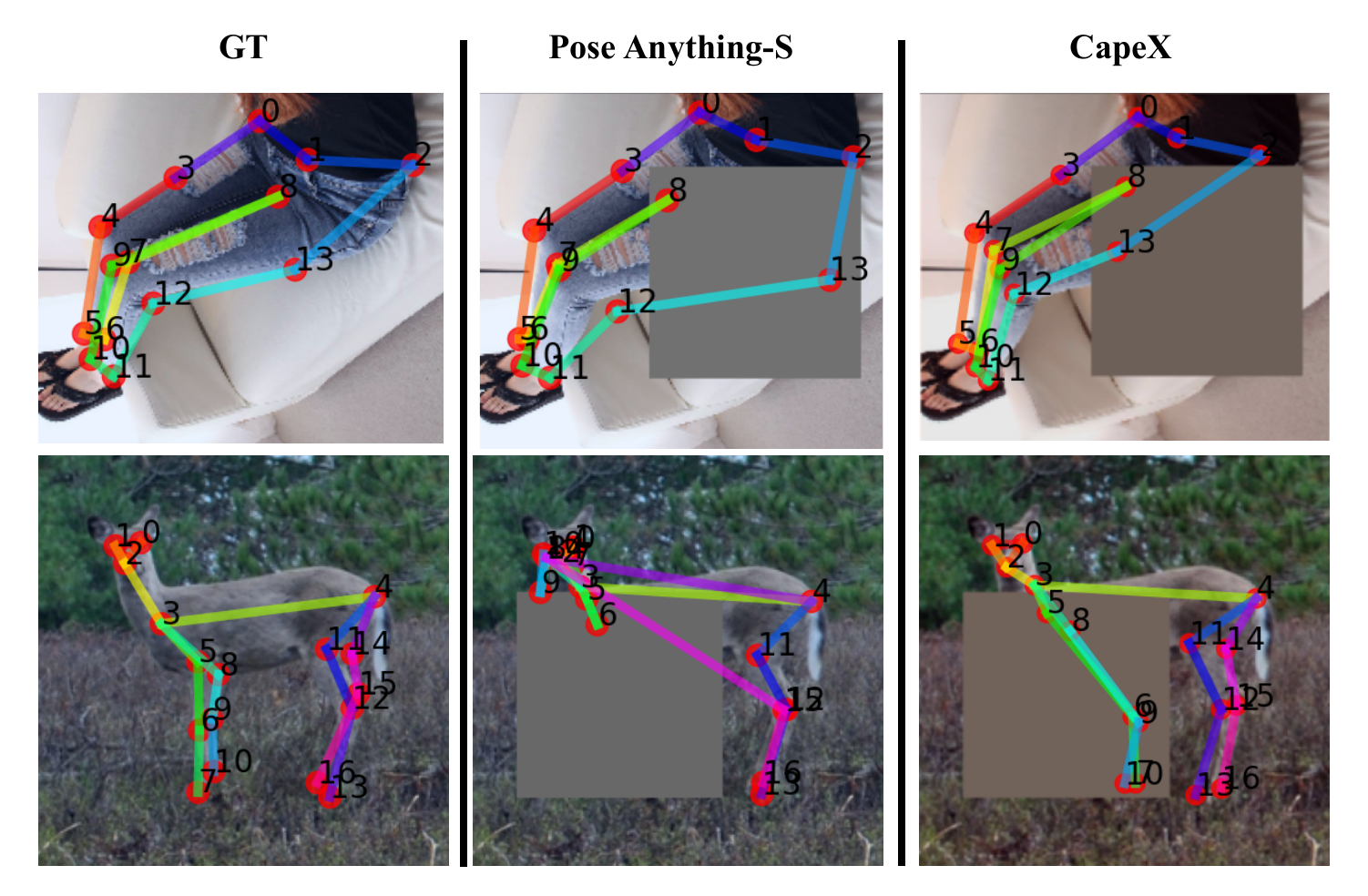}
    \caption{{\bf Comparison to Pose Anything-S:} Qualitative comparison between our model and Pose Anything-S on masked queries. CapeX does not require support images and can handle masked occlusions. Support images and text-graphs are not shown.
    }
    \label{fig:mask_example}
\end{figure}
\begin{figure}[h!]
    \centering
    \includegraphics[width=\linewidth]{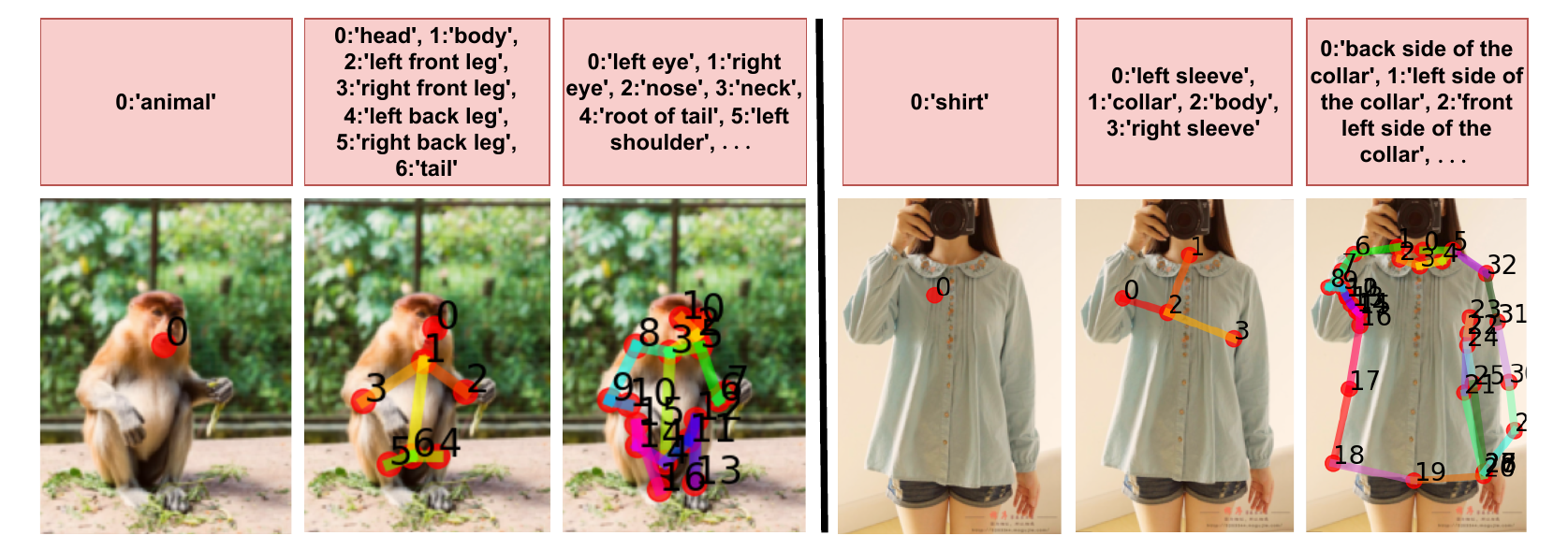}
    \caption{{\bf Text Abstractions:} Model performance over different levels of text-pose abstractions.}
    \label{fig:abstractions2}
\end{figure}

We present qualitative comparison between our support text-based framework and Pose Anything-S's support image-based framework in Figure \ref{fig:mask_example}. Our model demonstrates better performance due to the abstraction power of text-graphs, compared to the use of support image which may be more restrictive.

We include additional results of our model's performance on different levels of abstractions in Figure \ref{fig:abstractions2}.




\end{document}

%% file: Tables/benchmark.tex
\begin{table}
\centering
\caption{
{\bf MP-100 results:} PCK$_{0.2}$ performance under the 1-shot setting. Our approach outperforms other methods on average.}
\begin{adjustbox}{max width=\textwidth}
\begin{tabular}{c|ccccc|c}

\toprule
\textbf{Model} & Split 1 & Split 2 & Split 3 & Split 4 & Split 5 & Avg \\

\midrule
ProtoNet~\cite{snell2017prototypical} & 46.05 & 40.84 & 49.13 & 43.34 & 44.54 & 44.78 \\
MAML~\cite{finn2017model} & 68.14 & 54.72 & 64.19 & 63.24 & 57.20 & 61.50 \\
Fine-tuned~\cite{nakamura2019revisiting} & 70.60 & 57.04 & 66.06 & 65.00 & 59.20 & 63.58 \\
POMNet~\cite{xu2022pose} & 84.23 & 78.25 & 78.17 & 78.68 & 79.17 & 79.70 \\
CapeFormer~\cite{shi2023matching} &  89.45 & 84.88 & 83.59 & 83.53 & 85.09 & 85.31 \\
CapeFormer-S \cite{hirschorn2023pose} & 92.88 & 89.11 & 89.16 & 87.19 & 88.73 & 89.41 \\
Pose Anything-S \cite{hirschorn2023pose} & 93.66 & 90.42 & \textbf{89.79} & 88.68 & 89.61 & 90.43 \\
\midrule
\textbf{CapeX} & \textbf{95.62} & \textbf{90.94} & 88.95 & \textbf{89.43} & \textbf{92.57} & \textbf{91.50} \\

\bottomrule
\end{tabular}
\end{adjustbox}
\label{tab:mp100}
\end{table} 

%% file: Figures/same_query/same_query.tex
\begin{figure}[h!]
    \centering
    \begin{tabular}{ccccc}
        Support & Query & & Support & Query \\
        \includegraphics[height=0.13\textheight]{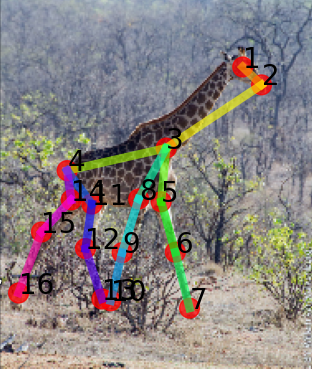} & 
        \includegraphics[height=0.13\textheight]{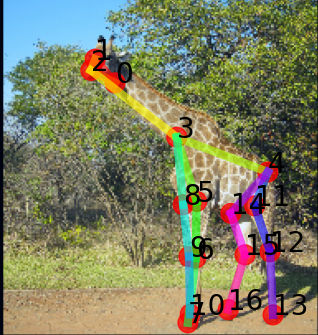} & &
        \includegraphics[height=0.13\textheight]{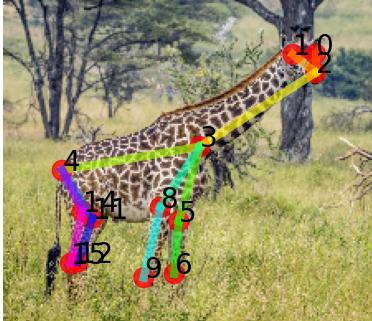} & 
        \includegraphics[height=0.13\textheight]{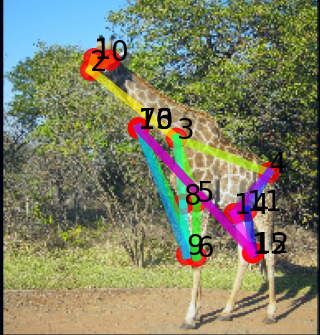} \\
        
        \includegraphics[height=0.13\textheight]{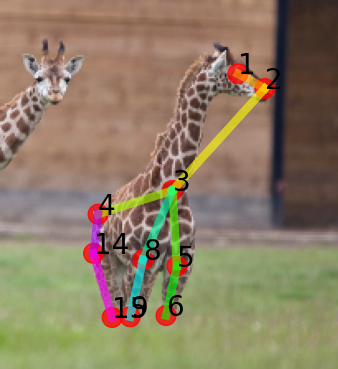} & 
        \includegraphics[height=0.13\textheight]{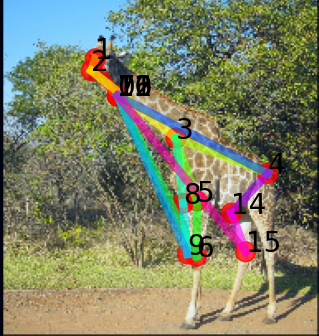} & &
        \includegraphics[height=0.13\textheight]{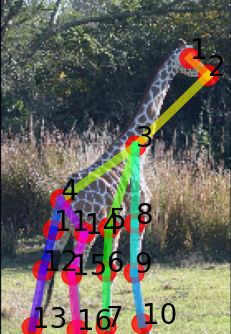} & 
        \includegraphics[height=0.13\textheight]{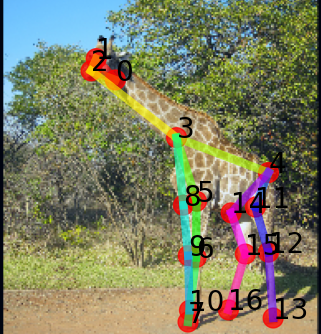} \\
    \end{tabular}
    \caption{\textbf{Visual Prompts Inconsistencies:} We show different results using Pose Anything model, for the same query image using different support images. Keypoints definitions and skeletons are the same. Using visual features impairs the ability to describe abstract semantic parts.
    }
  \label{fig:same_query}
\end{figure}

%% file: Tables/ablation_finetune.tex
\begin{table}[h!]
\caption{
{\bf Ablation experiments:} Tuning (T) VS. Freezing (F) the text backbone in model training, utilizing the graph transformer decoder or the original mlp transformer decoder. PCK$_{0.2}$ performance under 1-shot setting, with gte-base-v1.5 or CLIP ViT-B/32 as the text backbone.
}
\centering
\begin{adjustbox}{max width=\textwidth}
\begin{tabular}{c|ccccc|c}
\toprule
\textbf{Model} & Split 1 & Split 2 & Split 3 & Split 4 & Split 5 & Avg \\
\midrule
CapeX-CLIP-T-graph & 94.55 & 88.71 & 87.29 & 88.54 & 91.65 & 90.15  \\
CapeX-CLIP-F-graph & 95.17 & 88.88 & 87.72 & 88.24 & 91.81 & 90.37  \\
CapeX-gte-T-graph & \textbf{96.28} & 89.15 & \textbf{89.17} & 87.66 & 92.62 & 90.98  \\
CapeX-gte-F-mlp & 94.69 & 89.99 & 89.08 & \textbf{89.55} & \textbf{92.79} & 91.22 \\
\midrule
\textbf{CapeX-gte-F-graph} & 95.62 & \textbf{90.94} & 88.95 & 89.43 & 92.57 & \textbf{91.50} \\
\bottomrule
\end{tabular}
\end{adjustbox}
\label{tab:ablation_finetune}
\end{table} 